\documentclass[letterpaper]{article} 
\usepackage{aaai2026}  
\usepackage{times}  
\usepackage{helvet}  
\usepackage{courier}  
\usepackage[hyphens]{url}  
\usepackage{graphicx} 
\usepackage{natbib}  
\usepackage{caption} 
\usepackage{algorithm}
\usepackage{algorithmic}
\usepackage{amsmath}
\usepackage{amsthm}
\usepackage{amssymb}
\usepackage{mathrsfs}
\usepackage{MnSymbol}
\newtheorem{definition}{Definition}

\usepackage{multirow}

\usepackage{booktabs}
\usepackage[table]{xcolor}

\usepackage{booktabs}
\usepackage{multirow}

\usepackage{makecell}

\usepackage{newfloat}
\usepackage{listings}
\DeclareCaptionStyle{ruled}{labelfont=normalfont,labelsep=colon,strut=off} 
\floatstyle{ruled}
\newfloat{listing}{tb}{lst}{}
\floatname{listing}{Listing}
\newcommand{\ours}{HTG-GCL}

\title{HTG-GCL: Leveraging Hierarchical Topological Granularity from Cellular Complexes for Graph Contrastive Learning}
\author{
    Qirui Ji\textsuperscript{\rm 1 \rm 2} \equalcontrib,
    Bin Qin\textsuperscript{\rm 1 \rm 2} \equalcontrib,
    Yifan Jin\textsuperscript{\rm 1 \rm 2},
    Yunze Zhao\textsuperscript{\rm 1 \rm 2},\\
    Chuxiong Sun\textsuperscript{\rm 1},
    Changwen Zheng\textsuperscript{\rm 1},
    Jianwen Cao\textsuperscript{\rm 1},
    Jiangmeng Li\textsuperscript{\rm 1}\thanks{Corresponding author.}
}
\affiliations{
    \textsuperscript{\rm 1}National Key Laboratory of Space Integrated Information System, Institute of Software Chinese Academy of Sciences\\
    \textsuperscript{\rm 2}University of Chinese Academy of Sciences\\

    \{jiqirui2022, yifan2020, chuxiong2016, changwen, jiangmeng2019, jianwen\}@iscas.ac.cn, \\
    \{qinbin21, zhaoyunze19\}@mails.ucas.ac.cn
}

\usepackage{bibentry}

\begin{document}

\maketitle

\begin{abstract}
Graph contrastive learning (GCL) aims to learn discriminative semantic invariance by contrasting different views of the same graph that share critical topological patterns.
However, existing GCL approaches with structural augmentations often struggle to identify task-relevant topological structures, let alone adapt to the varying coarse-to-fine topological granularities required across different downstream tasks.
To remedy this issue, we introduce \textit{\textbf{H}ierarchical \textbf{T}opological \textbf{G}ranularity \textbf{G}raph \textbf{C}ontrastive \textbf{L}earning} (HTG-GCL), a novel framework that leverages transformations of the same graph to generate multi-scale ring-based cellular complexes, embodying the concept of topological granularity, thereby generating diverse topological views.
Recognizing that a certain granularity may contain misleading semantics, we propose a multi-granularity decoupled contrast and apply a granularity-specific weighting mechanism based on uncertainty estimation.
Comprehensive experiments on various benchmarks demonstrate the effectiveness of HTG-GCL, highlighting its superior performance in capturing meaningful graph representations through hierarchical topological information.
\end{abstract}

\begin{links}
    \link{Code}{https://github.com/ByronJi/HTG-GCL}
\end{links}

\section{Introduction}

Graph Contrastive Learning (GCL) has emerged as an effective paradigm for learning discriminative graph representations by contrasting multiple views of the same graph, offering significant promise in domains such as biochemical molecular classification and social network analysis \cite{gclsurvey}. 
Recent advances in structure-aware GCL methods \cite{graphcl,adgcl,joao,mega,rgcl,html} demonstrate the power of leveraging structural perturbations to generate informative views, enabling the model to effectively learn from unlabeled and non-Euclidean data.
However, as complex data from natural systems becomes increasingly prevalent, representing such data with simple pairwise relationships in graphs often falls short in capturing their inherent relational complexity.
To model these higher-order relational structures, more expressive mathematical objects beyond graphs, such as simplicial complexes and cellular complexes, have been introduced as generalized data representations \cite{madhu2024toposrl,cellclat}.
These topological abstractions allow for a richer encoding of multi-way interactions and geometric constraints, thus providing a principled way to incorporate higher-order topology into representation learning frameworks.

\begin{figure*}
    \centering
    \includegraphics[width=1\linewidth]{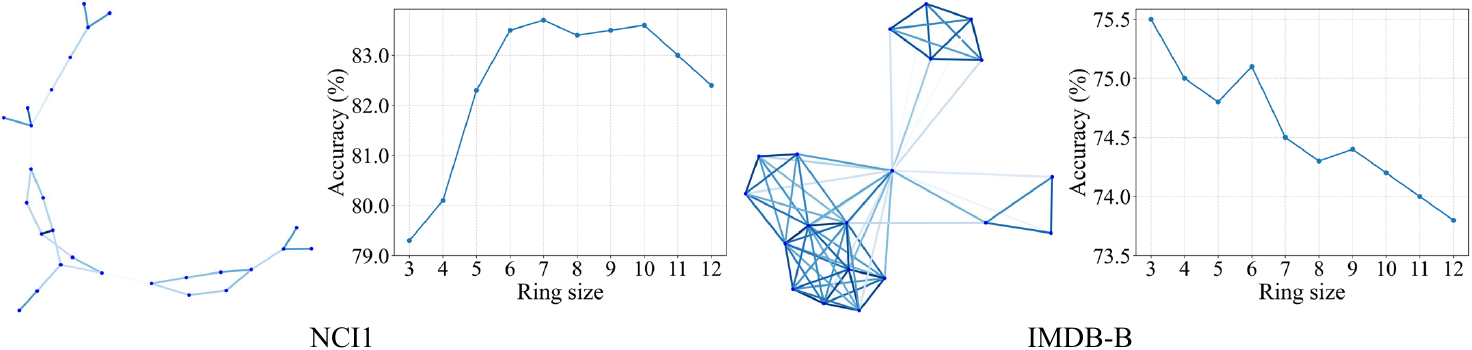}
    \caption{Visualization of graph structures \cite{graphvisual} and classification performance across different ring sizes on two TU datasets: NCI1 and IMDB-B. The results reveal that topological granularity significantly affects downstream performance, and indiscriminate inclusion of high-order structures may harm accuracy when they misalign with domain-specific semantics. For example, IMDB-B benefits from smaller rings (triangles), while NCI1 favors larger ring structures, highlighting the need for scenario-adaptive granularity modeling.}
    \label{fig:granularity-motivation}
\end{figure*}

However, current state-of-the-art GCL work \cite{cellclat} adopts a fixed ring size when constructing its cellular complexes, thereby capturing only a single level of topological \textit{granularity}.
Here, the \textit{granularity} of a cellular complex refers to the scale at which the graph’s topological structure is represented, determined by the maximum ring size $m$ used to attach 2-cells (see Section~\ref{sec:granularity-define} for details).
Such a static design assumes that a single topological resolution is universally effective across diverse domains—a premise that may not hold in practice.
Different real-world scenarios often depend on distinct types of topological patterns.
This observation is supported by our intuitive motivation experiments conducted in a supervised setting on the molecular dataset NCI1 and the social network dataset IMDB-B (Figure~\ref{fig:granularity-motivation}), which show that indiscriminately incorporating high-order structures may degrade performance when the extracted patterns do not align with the structural semantics of a specific domain.
For instance, social network analysis typically benefits from identifying triangle structures (i.e., 3-cliques), and thus operates effectively at a coarse topological granularity that captures small rings.
In contrast, molecular datasets, such as those used for toxicity prediction or mutagenicity detection, often require identifying larger and more complex ring structures, corresponding to a finer topological granularity.
This divergence reflects the need for scenario-dependent topological granularity.

\begin{figure}
    \centering
    \includegraphics[width=1\linewidth]{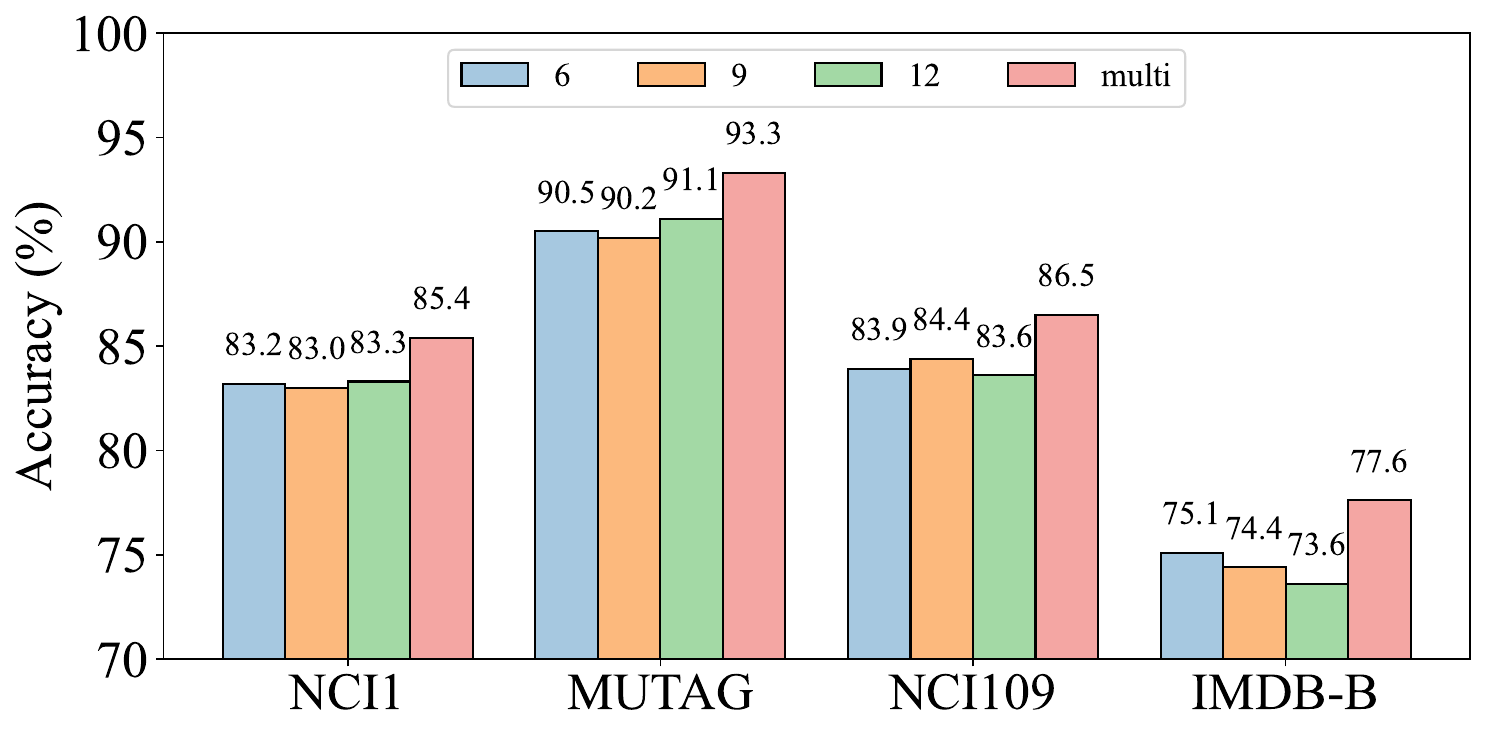}
    \caption{Performance comparison on four TU datasets under different single-ring granularities (6, 9, 12) and their multi-granularity integration. The multi-granularity setting consistently outperforms any single fixed granularity, demonstrating the complementary nature of hierarchical topological views.}
    \label{fig:ens-motivation}
\end{figure}

Yet in the context of GCL, the model lacks access to task-specific labels, making it difficult to determine which topological granularity is most relevant for downstream performance. A natural solution is to incorporate multiple levels of granularity—ranging from coarse to fine—so that the model can leverage a richer hierarchy of topological signals.
This motivates the following central question:

\textbf{\textit{``Can hierarchical multi-granularity topological modeling enhance downstream performance in GCL?''}}

With the above question in mind, we conduct an additional motivation experiment on four TU datasets (NCI1, MUTAG, NCI109, and IMDB-B) to explore the impact of multi-granularity topological modeling on downstream classification tasks.
As illustrated in Figure \ref{fig:ens-motivation}, we compare the classification accuracy achieved using cellular complexes constructed with different maximum ring sizes (6, 9, and 12), as well as a multi-granularity integration strategy that combines all three.
The results consistently show that the multi-granularity approach outperforms any single fixed granularity across all datasets, indicating that integrating hierarchical topological information provides complementary structural cues that enhance representation learning.


These findings motivate us to develop a principled approach that can effectively leverage hierarchical topological information in an self-supervised setting.
To this end, we introduce \textit{\textbf{H}ierarchical \textbf{T}opological \textbf{G}ranularity \textbf{G}raph \textbf{C}ontrastive \textbf{L}earning} (HTG-GCL), a novel framework that constructs diverse topological views by transforming the same graph into a series of ring-based cellular complexes with varying maximum ring sizes, thereby capturing multiple levels of topological granularity.
Motivated by the observation that different scenarios may favor different granularities, we recognize that not all topological views are equally informative for every downstream task.
To address this, we propose a \textit{multi-granularity decoupled contrastive learning strategy} (MGDC), where each view is projected into a granularity-specific embedding space, under the hypothesis that certain particular granularity may be uninformative or even misleading.
Moreover, since this hypothesis may not universally hold, we further incorporate an \textit{uncertainty-based weighting mechanism} to assess the reliability of each granularity-specific space and adaptively reweight their contributions during contrastive training.
Our contributions are summarized as follows:

\begin{itemize}
    \item We introduce the \textit{first} work that leverages cellular complexes to construct multi-granularity topological views for GCL, effectively capturing hierarchical topological information at different ring scales, which is a complementary perspective to traditional augmentations.
    \item We propose HTG-GCL, which generates multi-granularity topological views and incorporates MGDC, an uncertainty-weighted contrastive strategy over decoupled granularity-specific embedding spaces.
    \item We empirically validate the effectiveness of \ours\ through extensive experiments on multiple benchmarks.
\end{itemize}

\section{Related Work}

\subsection{Graph Contrastive Learning}
Recent advancements in graph contrastive learning have primarily focused on designing effective structural augmentation strategies and contrastive objectives to enhance graph-level representations. 
GraphCL \cite{graphcl} initiates this line of research by proposing four general augmentations to enforce consistency between perturbed graph views.
To improve augmentation quality, ADGCL \cite{adgcl} introduces adversarial strategies that avoid redundant semantics, while JOAO \cite{joao} tackles the problem of augmentation selection via a min-max bi-level optimization framework.
RGCL \cite{rgcl} incorporates rationale discovery to preserve meaningful structural patterns, and SimGRACE \cite{simgrace} avoids handcrafted augmentations altogether by perturbing the encoder with Gaussian noise.
HTML \cite{html} leverages knowledge distillation to align graph-level and subgraph-level views based on topological isomorphism.
TopoSRL \cite{madhu2024toposrl} introduces self-supervised learning on simplicial complexes with topology-preserving augmentations to capture higher-order interactions.
CellCLAT \cite{cellclat} goes beyond standard graphs by introducing cellular complexes, aiming to enrich higher-order structural information while addressing topological redundancy.
However, current methods lack the capacity to capture topological semantics at varying granularities. HTG-GCL addresses this by constructing multi-scale cellular complexes and utilizing granularity-decoupled contrastive learning.

\subsection{Higher-order Representations of Graphs}


Advancements in topological deep learning have shifted the focus from traditional graphs to more expressive structures, enabling the modeling of higher-order relationships~\cite{toposurvey2024}.
Simplicial complexes capture hierarchical connectivity by connecting nodes with edges, faces, and higher-dimensional simplices. Models such as SIN~\cite{sin} and SCNN~\cite{scnn} are well-suited for encoding these structures.
Cellular complexes further generalize simplicial complexes by allowing arbitrary shapes for faces and volumes, making them suitable for representing more intricate topological structures. Cellular Complex Neural Networks~\cite{hajij2020cell, bodnar2021weisfeiler2} have been proposed to effectively encode such data.


\section{Background}
\subsection{Cellular Complex}
\label{cellular complex}
A $d$-dimensional cellular complex $X$ is a topological space constructed inductively by a sequence of increasing $k$-skeleton: $X^{(-1)}=\varnothing\subset X^{(0)}\subset X^{(1)}\subset\cdots\subset X^{(d)}=X$. The construction~\cite{hansen2019toward} begins with a discrete set of points, called 0-cells, which form the 0-skeleton $X^{(0)}$. Then, for $k\geq 1$, the \textbf{$k$-skeleton} $X^{(k)}$ is formed by attaching \textbf{$k$-cells} $\sigma$ (topological spaces homeomorphic to the open $k$-dimensional ball $B^k=\left\{ x\in \mathbb{R}^k:\left\| x \right\|< 1 \right\}$) to the $(k-1)$-skeleton $X^{(k-1)}$. This attachment is guided by continuous map called \textbf{attaching map:} $\varphi_{\sigma}^k \colon\mathbb{S}^{k-1}\to X^{(k-1)}$, where $(k-1)$-sphere $\mathbb{S}^{k-1}$  is the \textbf{boundary} ($\partial \sigma \cong\mathbb{S}^{k-1}$) of the $k$-cell $\sigma$ and the index $\sigma$ ranges over all $k$-cells being attached. Finally, the entire cellular complex $X$ is the union of all its skeletons $X=\bigcup_{k=0}^{d}X^{(k)}$.
\subsubsection{Example 1 (Graph as a 1-dim cellular complex):} Graph $G=(V,E)$ is a 1-dim cellular complex $G=X^{(1)}$, where vertices $V$ are 0-cells that constitute the 0-skeleton $X^{(0)}$. Each edge $e = \{u,v\}\in E$ is regarded as a 1-cell $\sigma_e$ by attaching the endpoints of line segments to these vertices $u,v$. Thus, the 1-skeleton $X^{(1)}$ is precisely the graph $G$.
\subsubsection{Example 2 (2-dim cellular complex from cycles):} Starting from a graph $G$, we attach a 2-cell $\sigma_C$ (homeomorphic to a 2-dim disk) to an induced cycle $C = (v_1,v_2,\dots,v_n,v_1) \subseteq G$ along its boundary circle with the cyclic sequence of edges forming $C$. The resulting 2-skeleton $X^{(2)}$ is thus formed by filling in induced cycles of the original graph.

\subsection{Cellular Adjacencies}
We naturally extend the adjacency of nodes (0-cells) on graphs to hierarchical cross-dimensional higher-order interactions between edges (1-cells) and faces (2-cells). First, analogous to the incidence matrix on graphs, the incidence matrix between $(k-1)$-cells and $k$-cells is $\mathbf{B}_k \in \mathbb{R}^{N_{k-1}\times N_k}$, where $N_k$ denotes the number of $k$-cells in $X$. The $(i,j)$-th entry of $\mathbf{B}_k$ is 1 if the $i$-th $(k-1)$-cell is on the boundary of the $j$-th $k$-cell.

Based on the incidence matrix, we define four types of cellular adjacencies: 1) Boundary adjacent: $\mathbf{A}_{k}^{\mathcal{B}}=\mathbf{B}_{k+1}^T \in \mathbb{R}^{N_{k+1}\times N_k}$; 2) Co-boundary adjacent: $\mathbf{A}_{k}^{\mathcal{C}}=\mathbf{B}_{k} \in \mathbb{R}^{N_{k-1}\times N_k}$; 3) Upper adjacent: $\mathbf{A}_{k}^{\mathcal{U}}=\mathbf{B}_{k+1}\mathbf{B}_{k+1}^T \in \mathbb{R}^{N_{k}\times N_k}$; 4) Lower adjacent: $\mathbf{A}_{k}^{\mathcal{L} }=\mathbf{B}_{k}^T\mathbf{B}_{k} \in \mathbb{R}^{N_{k}\times N_k}$. The $k$-th Hodge Laplacian is $\mathbf{L}_k=\mathbf{A}_{k}^{\mathcal{U}}+\mathbf{A}_{k}^{\mathcal{L}}$. In particular, when $k=0$, $\mathbf{L}_0=\mathbf{B}_1\mathbf{B}_1^T=\mathbf{D}-\mathbf{A}$ is the usual graph Laplacian. The adjacency matrix of a graph is a special case of cellular adjacencies, namely the upper adjacent $\mathbf{A}_{0}^{\mathcal{U}}$ of nodes (0-cells). Finally, we use the notation $\mathcal{B}(\sigma)$, $\mathcal{C}(\sigma)$, $\mathcal{N}_{\mathcal{U}}(\sigma)$, and $\mathcal{N}_{\mathcal{L}}(\sigma)$ to denote the boundary, co-boundary, upper and lower neighbors of cell $\sigma$, respectively.

\begin{figure*}
    \centering
    \includegraphics[width=1\linewidth]{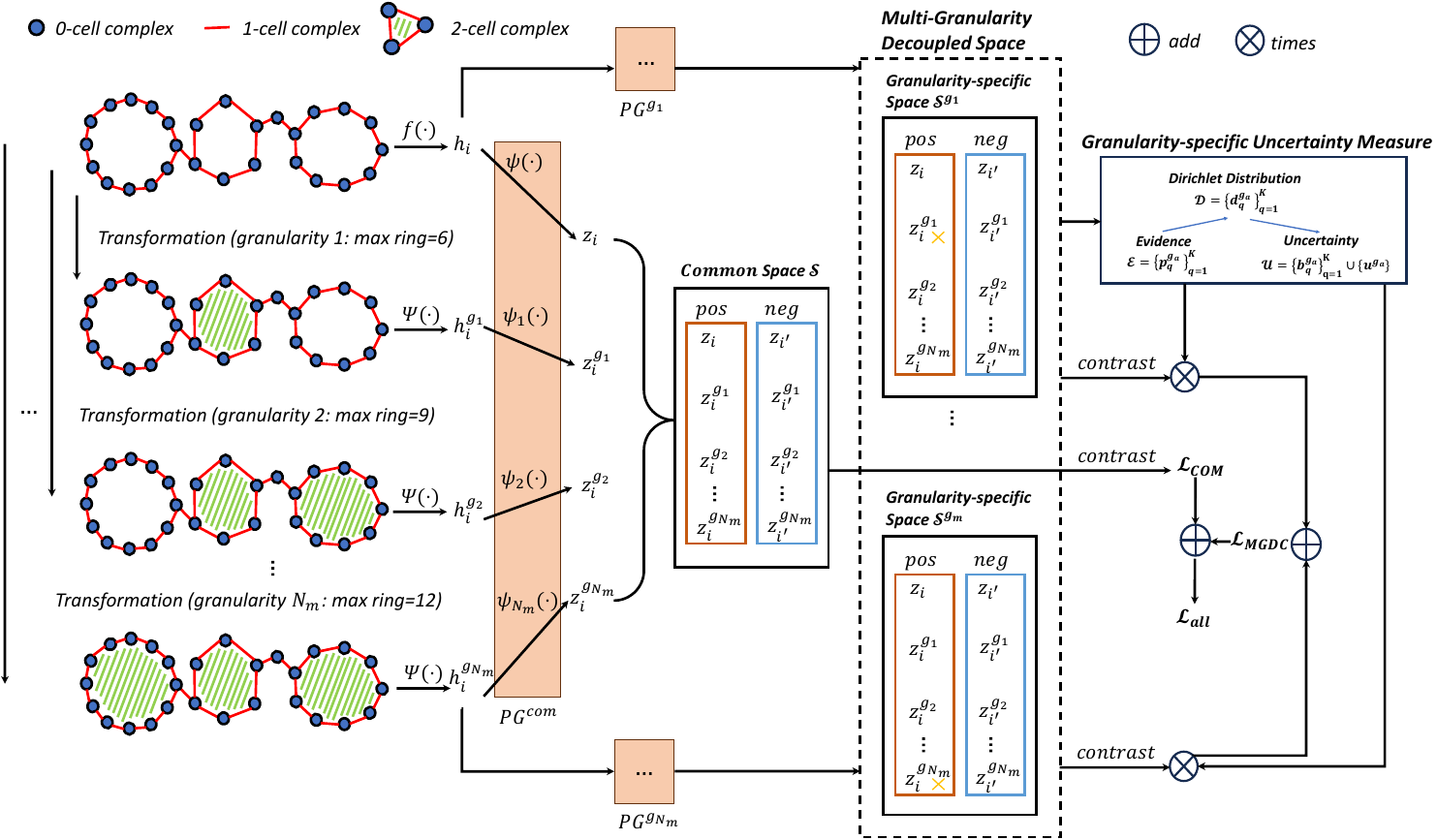}
    \caption{The framework of HTG-GCL.}
    \label{fig:framework}
\end{figure*}

\subsection{Cellular Complex Neural Networks}
Let the initial features of the $k$-cells in a cellular complex $X$ be denoted as $\mathbf{X}_k\in \mathbb{R}^{N_k \times F_k}$. Cellular Complex Neural Networks (CCNNs) encode these $k$-cells and learn their representations by aggregating messages from four types of neighbors. Specifically, the message-passing paradigm~\cite{hajij2020cell,bodnar2021weisfeiler2} for learning the $l$-th embedding $\mathbf{Z}_k^{(l)}$ of $k$-cells in a CCNN can be unified as follows:
\begin{equation}
    \begin{aligned}
        &\mathbf{Z}_k^{(l)} = \psi \mathopen{[} \bigoplus \mathopen{(}
        \mathbf{A}_{k-1}^{\mathcal{B}} \mathbf{Z}_{k-1}^{(l-1)} \mathbf{W}^{(l-1)}_{\mathcal{B}},\ 
        \mathbf{A}_{k+1}^{\mathcal{C}} \mathbf{Z}_{k+1}^{(l-1)} \mathbf{W}^{(l-1)}_{\mathcal{C}}, \\
        & \mathbf{A}_{k}^{\mathcal{U}} \mathbf{Z}_{k}^{(l-1)} \mathbf{W}^{(l-1)}_{\mathcal{U}},\ 
        \mathbf{A}_{k}^{\mathcal{L}} \mathbf{Z}_{k}^{(l-1)} \mathbf{W}^{(l-1)}_{\mathcal{L}},\ 
        \mathbf{Z}_{k}^{(l-1)} \mathbf{W}^{(l-1)}
        \mathclose{)} \mathclose{]},
    \end{aligned}
\end{equation}
where $\bigoplus$ denotes a differentiable, permutation-invariant aggregation function, \textit{e.g.}, sum, mean, or concatenation, and $\psi$ is a nonlinear updating function such as ReLU. The initial embedding is defined as $\mathbf{Z}_k^{(0)}=\mathbf{X}_k$. After $L$ layers of message passing, the embeddings of 0-cells (nodes), 1-cells (edges), and 2-cells (faces) are denoted as $\mathbf{Z}_0^{(L)}$, $\mathbf{Z}_1^{(L)}$, and $\mathbf{Z}_2^{(L)}$, respectively. The final representation $\mathbf{Z}_{X}$ of the cellular complex $X$ is then given by:
\begin{equation}
    \mathbf{Z}_{X} =\mathrm{READOUT}\left ( \sum_{k=0}^{2}\text{MLP}_k\,(\phi_k(\mathbf{Z}_k^{(L)})) \right ), 
\end{equation}
where $\phi_k:\mathbb{R}^{N_k\times H_k}\!\longrightarrow\!\mathbb{R}^{1\times H_k}$ is a dimension‑specific local pooling function (e.g., sum or mean). $\mathrm{READOUT}$ is a differentiable global pooling function such as MLP that gets the final graph‑level representation $\mathbf{Z}_{X}\in\mathbb{R}^{H}$.

\section{Methodology}
In this paper, we focus on developing a novel GCL learning framework which is shown in Figure \ref{fig:framework}.

\subsection{Ring‑based Graph‑to‑Complex Preprocessing}
\label{sec:granularity-define}
We model and represent hierarchical topological granularity at the data level, where transforming graphs to cellular complexes constitutes a one-time preprocessing step prior to training. As mentioned in the background Section \ref{cellular complex}, there exist multiple approaches for transforming graphs into cellular complexes. In this work, we attach 2-cells (2-dimensional disks) to all rings containing at most $m$ edges in the graph, thereby obtaining a nested sequence of cellular complexes.

\begin{definition}[Ring-based Cellular Complexes Transformation]
    Given a finite undirected graph $G=(V,E)$, let 
    $$\mathcal{R}_{\le m}(G)=\bigl\{\,R\subseteq G\mid R\text{ is a ring and }|E(R)|\le m\,\bigr\}$$
    denote the set of all rings in $G$ whose length is at most $m$. Define a 2‑dimensional cellular complex $X_{m}(G)$ by taking $G$ as the 1‑skeleton and attaching one 2‑cell along the edges of every ring $R \in\mathcal{R}_{\le m}(G)$. The transformation $\mathcal{T}_{\mathrm{Ring}}\colon G\;\mapsto\;\{X_{m}(G)\}_{m\in\mathbb{N}_{\ge 3}}$ is called the \textbf{Ring‑based Cellular Complexes Transformation} of $G$.
\end{definition}

The cellualr complex sequence $\{X_m(G)\}_m $ is hierarchical. Although they all share the same 1-skeleton as graph $G $, they contain different topological information. As $m$ increases, $X_m(G)$ provides increasingly fine-grained characterizations of the structure of graph $G$. However, excessive focus on fine-grained features may be sub‑optimal on tasks that naturally target lower‑order motifs.

\begin{definition}[Hierarchical Topological Granularity]
     Given the cellular complex sequence $\{X_m(G)\}_m$, we say that $X_n(G)$ has \textbf{finer topological granularity} than $X_m(G)$, denoted by $X_n(G)\succeq X_m(G)$, if $n>m$. Formally, increasing the granularity from $X_m(G)$ to $X_n(G)$ introduces additional topological features captured by larger rings, thus providing a richer and more detailed characterization of higher-order structures in graph $G$. Conversely, smaller rings correspond to \textbf{coarser topological granularity}, capturing only basic or low-order structural motifs. 
\end{definition}

\subsection{Multi-scale Topological Granularity Decoupled Contrastive Learning}




In our method, we transform the original graph into $N_m$ cellular complexes of different topological granularities.
Given a graph $G_i$, we apply ring-based transformations ${\mathcal{T}_{g_a}}({a \in \{1,...,N_m\}})$ to generate a set of cellular complexes $\{X_{g_a}(G_i)\}_{g_a}$, where $g$ is the set of topological granularities. Each $\{X_{g_a}(G_i)\}_{g_a}$ is encoded by a shared CCNN encoder $\Psi(\cdot)$ to obtain embedding $h_i^{g_a}$, while the original graph is encoded by a GIN~\cite{xu2018powerful} encoder $f(\cdot)$ to produce $h_i$.
Note that in practice, we set $N_m=3$, corresponding to maximum ring sizes of 6, 9, and 12, respectively.

\subsubsection{Common Space Contrast}

In Common space, all these embeddings are passed through a projection group $PG^{com}$ and mapped into a common embedding space $\mathcal{S}$, where we assume that all granularities offer informative semantic cues. We denote the resulting projected feature of the $j$-th view of graph $G_i$ as $z_i^{g_j}$, where the 0-th (GIN-based) view is denoted by $z_i$.
We then optimize a contrastive objective \cite{cmc,mvgrl,DBLP:journals/tkde/LiQZSRWX23} in this space, which encourages consistency across different views of the same graph while enforcing distinction from other graphs. The loss is formally defined as:

\begin{equation}\label{eq:comloss}\small
\mathcal{L}_{COM} = - \sum_i \sum_{\substack{j,k \in [0,N_m]\\ j \ne k}} 
\log \left[
\frac{
s(i,j,k)
}{
s(i,j,k)
+ \sum_{\substack{i' \in [1,n], i' \ne i\\ j',k' \in [0,N_m]}} s(i',j',k')
}
\right]
\end{equation}

\noindent where $s(i,j,k) = \exp\left( c(z_i^j, z_i^k) / \tau \right)$ defines the similarity between a positive pair, $c(\cdot,\cdot)$ denotes cosine similarity, and $\tau$ is a temperature scaling factor. The term $s(i',j',k')$ represents the similarity between negative pairs drawn from different samples.


\begin{table*}[h]
	\begin{center}
			\begin{tabular}{c|cccc|cc|c}
				\hline
                \hline
				\text{Dataset} & \text{NCI1} & \text{PROTEINS} & \text{MUTAG} 
& \text{NCI109} & \text{IMDB-B} 
& \text{IMDB-M} & A.R. $\downarrow$ \\
				\hline
                    \hline
                    \text{node2vec}  & 54.9 $\pm$ 1.6 & 57.5 $\pm$ 3.6 & 72.6 $\pm$ 10.0 
& -& - 
& - & 13.0 \\
                    \text{sub2vec}  & 52.8 $\pm$ 1.5 & 53.0 $\pm$ 5.6 & 61.1 $\pm$ 15.8 
& -& 55.3 $\pm$ 1.5 
& - & 13.8 \\
                    \text{graph2vec}  & 73.2 $\pm$ 1.8 & 73.3 $\pm$ 2.0 & 83.2 $\pm$ 9.3 
& -& 71.1 $\pm$ 0.5 
& -& 11.0 \\
                    \text{InfoGraph}  & 76.2 $\pm$ 1.0 & 74.4 $\pm$ 0.3 & 89.0 $\pm$ 1.1 
& 76.2 $\pm$ 1.3 & 73.0 $\pm$ 0.9 
&  48.1 $\pm$ 0.3 & 7.5 \\
                    \text{GraphCL}  & 77.9 $\pm$ 0.4 & 74.4 $\pm$ 0.5 & 86.8 $\pm$ 1.3 
& 78.1 $\pm$ 0.4 & 71.2 $\pm$ 0.4 
& 48.9 $\pm$ 0.3 & 8.0 \\
                    \text{ADGCL}  & 73.9 $\pm$ 0.8 & 73.3 $\pm$ 0.5 & 88.7 $\pm$ 1.9 
& 72.4 $\pm$ 0.4& 70.2 $\pm$ 0.7 
& 48.1 $\pm$ 0.4& 10.2 \\
                    \text{JOAO}  & 78.1 $\pm$ 0.5 & 74.6 $\pm$ 0.4 & 87.4 $\pm$ 1.0 
& 77.2 $\pm$ 0.6 & 70.2 $\pm$ 3.1 
& 48.9 $\pm$ 1.2 & 8.5 \\
                    \text{JOAOv2}  & 78.4 $\pm$ 0.5 & 74.1 $\pm$ 1.1 & 87.7 $\pm$ 0.8 
& 78.2 $\pm$ 0.8 & 70.8 $\pm$ 0.3 
& 49.2 $\pm$ 0.9 & 7.0 \\
                    \text{RGCL}  & 78.1 $\pm$ 1.0 & 75.0 $\pm$ 0.4 & 87.7 $\pm$ 1.0 
& 77.7 $\pm$ 0.3& 71.9 $\pm$ 0.9 
& 49.3 $\pm$ 0.4 & 6.2 \\
                    \text{SimGRACE}  & 79.1 $\pm$ 0.4 & 75.3 $\pm$ 0.1 & 89.0 $\pm$ 1.3 
& 78.4 $\pm$ 0.4 & 71.3 $\pm$ 0.8 & 49.1 $\pm$ 0.8 &  4.3 \\
                    \text{HTML} & 78.2 $\pm$ 0.7 & 75.0 $\pm$ 0.3 & 88.9 $\pm$ 0.8 & 77.9 $\pm$ 0.2 & 71.7 $\pm$ 0.4 & 48.9 $\pm$ 0.6 & 6.0 \\
                   \text{DRGCL} & 78.7 $\pm$ 0.4 & 75.2 $\pm$ 0.6 & 89.5 $\pm$ 0.6 & 77.8 $\pm$ 0.4 & 72.0 $\pm$ 0.5 & 49.5 $\pm$ 0.5 & 4.2 \\
                     \text{CellCLAT} & \underline{79.4 $\pm$ 0.2} & \underline{75.7 $\pm$ 0.1} & \underline{89.7 $\pm$ 0.3} & \underline{78.9 $\pm$ 0.4} & \underline{73.4 $\pm$ 0.1} & \underline{50.6 $\pm$ 0.2} & \underline{2.0} \\
        
  \hline

 \text{\ours}& \textbf{80.7 $\pm$ 0.4}\textsuperscript{***} & \textbf{76.2 $\pm$ 0.5}\textsuperscript{**} & \textbf{90.4 $\pm$ 0.5}\textsuperscript{**} & \textbf{79.7 $\pm$ 0.4}\textsuperscript{***} & \textbf{74.0 $\pm$ 0.4}\textsuperscript{***} & \textbf{51.5 $\pm$ 0.5}\textsuperscript{***} & \textbf{1.0}\\
 \hline
 \hline
			\end{tabular}
	\end{center}
 \caption{Unsupervised representation learning classification accuracy (\%) on TU datasets. A.R denotes the average rank of the results. The best results are highlighted in \textbf{bold}, and the second best results are highlighted with \underline{underline}. Statistical significance is determined via a one-sided two-sample t-test. *, **, and *** indicate p-values less than 0.1, 0.05, and 0.01, denoting marginally, statistically, and strongly significant results, respectively.}
	\label{tab:htggcl_unsupervised_learning}
\end{table*}

\begin{table*}[t]
	\begin{center}
			\begin{tabular}{c|ccc|cc|c}
				\hline
                \hline
				\text{Dataset} & \text{NCI1} & \text{PROTEINS}
& \text{NCI109} & \text{IMDB-B} 
& \text{IMDB-M} & A.R. $\downarrow$ \\
				\hline
                    \hline
                    \text{InfoGraph}  & 68.9 $\pm$ 0.8 & 69.5 $\pm$ 0.9    
& 67.8 $\pm$ 0.7 & 68.5 $\pm$ 0.6 
&  43.8 $\pm$ 0.2 & 8.2 \\
                    \text{GraphCL}  & 69.6 $\pm$ 0.3 & 69.7 $\pm$ 0.1   
& 69.2 $\pm$ 0.3 & 67.8 $\pm$ 1.3 
& 43.9 $\pm$ 0.3 & 6.0 \\
                    \text{ADGCL}  & 65.6 $\pm$ 0.6 & 68.5 $\pm$ 2.0 
& 65.7 $\pm$ 0.8 & 67.7 $\pm$ 0.9 
& 41.9 $\pm$ 0.9& 10.8 \\
                    \text{JOAO}  & 69.1 $\pm$ 0.1 & 70.9 $\pm$ 2.0  
& 69.2 $\pm$ 0.3 & 67.6 $\pm$ 0.8 
& 43.9 $\pm$ 1.0 & 6.6 \\
                    \text{JOAOv2}  & 69.1 $\pm$ 0.4 & 70.1 $\pm$ 1.4  
& 69.2 $\pm$ 0.2 & 68.3 $\pm$ 1.0 
& 43.6 $\pm$ 1.1 & 6.8 \\
                    \text{RGCL}  & 70.2 $\pm$ 0.7 & 71.2 $\pm$ 0.9 
& 69.1 $\pm$ 0.3 & \underline{68.9 $\pm$ 0.3} 
& \underline{44.5 $\pm$ 0.9} & 3.4 \\
                    \text{SimGRACE}  & 69.3 $\pm$ 0.3 & 71.3 $\pm$ 0.7  
& 69.0 $\pm$ 0.2 & 68.6 $\pm$ 0.7 & 44.2 $\pm$ 0.6 & 4.6 \\
                    \text{HTML} & 69.4 $\pm$ 0.1 & 69.1 $\pm$ 1.5 & 68.9 $\pm$ 0.1 & 68.4 $\pm$ 0.3 & 43.2 $\pm$ 0.3 & 7.8 \\
                    \text{DRGCL} & 69.3 $\pm$ 0.2 & 70.3 $\pm$ 0.4 &  68.4 $\pm$ 0.2 & 68.0 $\pm$ 0.6 & 44.3 $\pm$ 0.5 & 6.4 \\
                     \text{CellCLAT} & \underline{70.4 $\pm$ 0.5} & \underline{71.8 $\pm$ 0.1}  & \underline{69.5 $\pm$ 0.3} & 68.5 $\pm$ 0.4 & 44.0 $\pm$ 0.3 & \underline{3.0} \\
        
  \hline
 \text{\ours}& \textbf{70.8 $\pm$ 0.6}\textsuperscript{*} & \textbf{72.3 $\pm$ 0.4}\textsuperscript{**}  & \textbf{70.6 $\pm$ 0.4}\textsuperscript{***} & \textbf{69.1 $\pm$ 0.5}\textsuperscript{*} & \textbf{45.2 $\pm$ 0.7}\textsuperscript{*} & \textbf{1.0} \\
 \hline
 \hline
			\end{tabular}
	\end{center}
 \caption{Semi-supervised representation learning classification accuracy (\%) on TU datasets.}
	\label{tab:htggcl_semisupervised_learning}
\end{table*}

\subsubsection{Multi-Granularity Decoupled Contrast}

Although the common embedding space $\mathcal{S}$ enables joint contrastive learning across multiple granularities, not all topological views are equally informative, i.e., certain granularities may contain misleading patterns for the downstream task. To address this, we introduce a set of projection heads $PG^{g_a}$ for each granularity $g_a$, which map the multi-view representations into granularity-specific decoupled spaces $\{\mathcal{S}^{g_a}\}_{a=1}^{N_m}$.
In each decoupled space $\mathcal{S}^{g_a}$, we treat the projection $z_i$ of the original graph $G_i$ as the anchor. The projected features $\{z_i^{g_{a'}}\}_{a' \ne a}^{N_m}$, obtained from other granularities, are considered as positive samples, while representations from different graphs serve as negatives. 
Notably, we exclude the view $a' = a$ from being used as a positive, as this space-specific assumption posits that the $g_a$-view may not contain reliable complementary information.
We then apply contrastive learning in space $\mathcal{S}^{g_a}$:
\begin{equation}\label{eq:decouple_loss}\small
\mathcal{L}_{\mathcal{S}^{g_a}} = - \sum_i \sum_{\substack{j=0,k \ne a,\\ k \in [0,N_m]}} 
\log \left[
\frac{
s(i,j,k)
}{
s(i,j,k)
+ \sum_{\substack{i' \in [1,n], i' \ne i\\ j',k' \in [0,N_m]}} s(i',j',k')
}
\right]
\end{equation}
By aggregating the contrastive losses across all granularity-specific decoupled spaces, we obtain the MGDC objective:
\begin{equation}
\mathcal{L}_{MGDC} = \sum_{a=1}^{N_m} \Omega_{g_a} \cdot \mathcal{L}_{\mathcal{S}^{g_a}},
\end{equation}
where $\Omega_{g_a}$ is the trustworthiness weight of granularity $g_a$, reflecting the reliability of its representation space, which is detailed in the following part.


\subsubsection{Granularity-specific Weighting}

We quantify this trustworthiness by estimating the uncertainty of the learned representations in each space. Intuitively, if a decoupled space induces representations with high clustering uncertainty, it is likely that the corresponding granularity fails to capture consistent semantic patterns. To model this, we adopt a Dirichlet distribution-based formulation \cite{subjectivelogic,tmc}.
For each sample, we first compute the clustering probability $p_{iq}^{g_a}$ of assigning the feature $z_i^{g_a}$ to cluster $q$, using the Student’s t-distribution:
\begin{equation}\label{eq:student-t}
p_{iq}^{g_a} = \frac{(1 + \|z_i^{g_a} - c_q^{g_a}\|^2 / \mu)^{-\frac{\mu+1}{2}}}{\sum_{q'} (1 + \|z_i^{g_a} - c_{q'}^{g_a}\|^2 / \mu)^{-\frac{\mu+1}{2}}},
\end{equation}
where $\mu$ is a degrees-of-freedom parameter, and $c_q^{g_a}$ is the centroid of cluster $q$ in space $\mathcal{S}^{g_a}$, formalized as:
\begin{equation}
c_q^{g_a} =
\begin{cases}
\frac{1}{N_m+1} \sum_{j=0}^{N_m} z_q^{j}, & \text{in common space } \mathcal{S}, \\
\frac{1}{N_m} \sum_{j=0, j \ne a}^{N_m} z_q^{j}, & \text{in decoupled space } \mathcal{S}^{g_a},
\end{cases}
\end{equation}
We interpret the clustering distribution $p_{q}^{g_a}$ as evidence and map it to a Dirichlet distribution with parameters $d_q^{g_a} = p_{q}^{g_a} + 1$. The overall uncertainty mass $u^{g_a}$ and belief masses $b_q^{g_a}$ are computed as:
\begin{equation}
b_q^{g_a} = \frac{p_{q}^{g_a}}{\sum_{q'} d_{q'}^{g_a}}, \quad u^{g_a} = \frac{K}{\sum_{q'} d_{q'}^{g_a}},
\end{equation}
where $K$ is the number of clusters (i.e. batchsize).
A lower Dirichlet uncertainty indicates a more confident and compact clustering of features, implying a higher trustworthiness of the corresponding granularity.
Finally, we aggregate the instance-wise uncertainty into the trustworthiness score for granularity $g_a$ as:
\begin{equation}
\Omega_{g_a} = \text{sgm}\left( \sum_{i=1}^n \sum_{q=1}^K \mathbb{I}_{iq} \cdot \left[ \text{dgm}(d_q^{g_a}) - \text{dgm}\left( \sum_{q'} d_{q'}^{g_a} \right) \right] \right),
\end{equation}
where $\text{dgm}(\cdot)$ is the digamma function, $\mathbb{I}_{iq}$ indicates the cluster assignment, and $\text{sgm}(\cdot)$ is the sigmoid function to normalize the result.
This adaptive mechanism enables the model to emphasize the contributions of reliable granularities while suppressing the influence of potentially noisy or uninformative ones.

\subsubsection{Training Objective}

To integrate the complementary strengths of both the common and granularity-specific contrastive views, we formulate the overall training objective as a weighted combination of the joint contrastive loss in the common space $\mathcal{S}$ and the granularity-aware decoupled contrastive losses $\mathcal{L}_{\mathcal{S}^{g_a}}$:
\begin{equation}
\mathcal{L}_{\text{all}} = \alpha \cdot \mathcal{L}_{COM} + \frac{\beta}{N_m} \sum_{a \in \{ 1,...,N_m \} } \Omega_{g_a} \cdot \mathcal{L}_{\mathcal{S}^{g_a}},
\end{equation}
where $\alpha$ and $\beta$ are balancing hyper-parameters that control the relative contributions of common contrast and decoupled contrast.
While the contrastive learning framework encourages the model to align and discriminate across multiple topological granularities during training, we further leverage this multi-granularity information at inference time. Specifically, we concatenate the embeddings after the encoders from all granularity-specific views $\{h_i^{g_a}\}_{a \in \{ 1,...,N_m \} }$ and the graph representation $h_i$ as the final graph-level embedding:
\begin{equation}
h_i^{\text{final}} = \text{CONCAT}\left(h_i, h_i^{g_1}, h_i^{g_2}, \dots, h_i^{g_{N_m}}\right).
\end{equation}
$h_i^{\text{final}}$ is then used for downstream classification, allowing the model to fully exploit the hierarchical topological granularity encoded by the cellular complex transformations.

\section{Experiments}

\subsection{Experimental Setup}

\subsubsection{Benchmarks}

We evaluate the performance of \ours\ on six widely adopted benchmarks from the TU collection \cite{tudataset}, including four bioinformatics datasets (NCI1, PROTEINS, MUTAG, NCI109) and two social network datasets (IMDB-B, IMDB-M). For unsupervised evaluation, all datasets are used, while in the semi-supervised setting, MUTAG is excluded due to its small size and severe class imbalance under k-fold cross-validation.

In the unsupervised setting, \ours\ is trained on the full dataset without label supervision. The learned graph-level representations are evaluated via a downstream classification task using an SVM classifier with 10-fold cross-validation. We report the mean and standard deviation of classification accuracy across five different random seeds.

For the semi-supervised evaluation, the same training procedure is used, but only 10\% of the labeled data is employed for SVM training. To leverage the remaining unlabeled data, we adopt a pseudo-labeling strategy: a classifier is first trained on the small labeled subset and then used to generate pseudo-labels for the unlabeled samples. A final model is trained using both labeled and pseudo-labeled data.


\subsubsection{Compared Baselines}
\begin{table*}[t]
	\begin{center}
			\begin{tabular}{c|cccc}
				\hline
                \hline
				\text{Dataset} & \text{NCI1} & \text{MUTAG}   
& \text{IMDB-B} 
& \text{IMDB-M} \\
				\hline
                    \hline
                    \text{One granularity (ring 6)}  &  79.3 $\pm$ 0.3& 89.7 $\pm$ 0.3 &  73.2 $\pm$ 0.3&  50.5 $\pm$ 0.3 \\
                     \text{One granularity (ring 9)} &  79.5 $\pm$ 0.4& 89.3 $\pm$ 0.5  
&  73.3 $\pm$ 0.2&  50.8 $\pm$ 0.6\\
                     \text{One granularity (ring 12)} &  79.5 $\pm$ 0.3& 89.4 $\pm$ 0.4    
&  72.8 $\pm$ 0.5&  50.7 $\pm$ 0.5\\
                     \hline
                     \text{Two granularities (ring 6, 9)}  & 79.9 $\pm$ 0.4& 89.9 $\pm$ 0.3&  73.4 $\pm$ 0.2&  51.0 $\pm$ 0.3\\
                    \text{Two granularities (ring 6,12)}  & 80.1 $\pm$ 0.6& 89.5 $\pm$ 0.5&  73.2 $\pm$ 0.4&  51.1  $\pm$ 0.4\\
                     \text{Two granularities (ring 9, 12)} & 80.1 $\pm$  0.4& 89.7 $\pm$ 0.6& 73.4 $\pm$ 0.3& 51.1 $\pm$ 0.3\\
                     \text{Two granularities (ring 6, 9) w MGDC}  & 80.3 $\pm$ 0.2& 90.1 $\pm$ 0.3 &  73.7 $\pm$ 0.2&  51.2 $\pm$ 0.3\\
                    \text{Two granularities (ring 6,12) w MGDC}  & 80.4 $\pm$ 0.4& 89.8 $\pm$ 0.4&  73.5 $\pm$ 0.2&  51.4 $\pm$ 0.4\\
                     \text{Two granularities (ring 9, 12) w MGDC} & 79.8 $\pm$ 0.4& 89.6 $\pm$ 0.3& 73.7 $\pm$ 0.3 & 51.4 $\pm$ 0.3\\
                     \hline
 \text{\ours\ w/o MGDC} & 79.7 $\pm$ 0.2 & 89.8 $\pm$ 0.4&  73.4 $\pm$ 0.3 & 50.9 $\pm$ 0.3\\
 \text{\ours} &  \textbf{80.7 $\pm$ 0.4} & \textbf{90.4 $\pm$ 0.5} & \textbf{74.0 $\pm$ 0.4} & \textbf{51.5 $\pm$ 0.5} \\
 \hline
 \hline
			\end{tabular}
	\end{center}
 \caption{Ablation studies on the unsupervised settings. w and
  w/o denote with and without, respectively.}
	\label{tab:ablation study}
\end{table*}

\begin{figure*}
    \centering
    \includegraphics[width=0.9\linewidth]{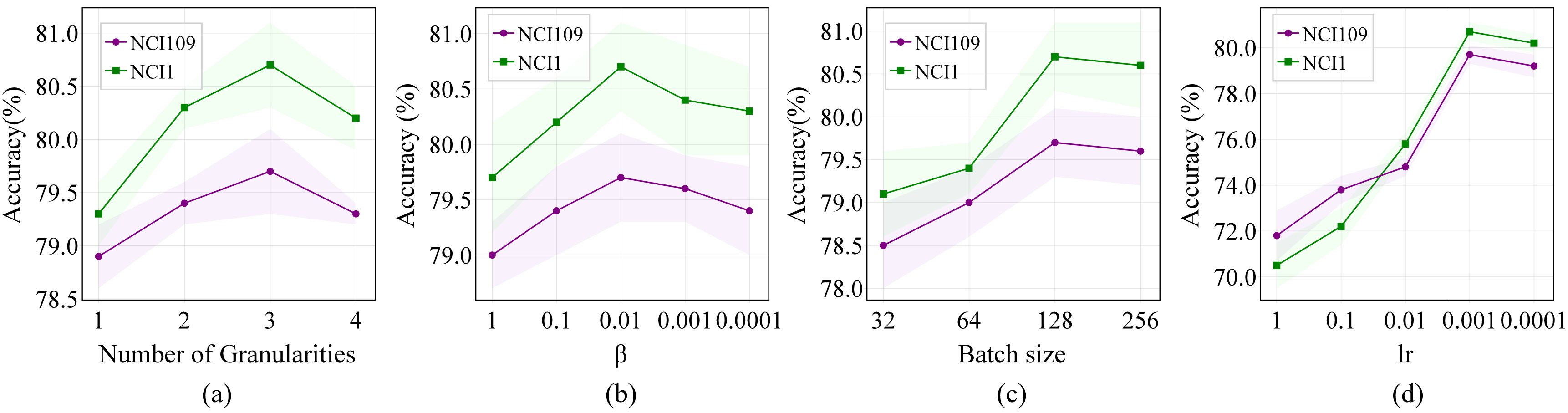}
    \caption{Hyper-parameter analysis.}
    \label{fig:hyper-main}
\end{figure*}

In the unsupervised scenario, we compare \ours\ against fourteen representative baselines: Node2Vec \cite{node2vec}, Sub2Vec \cite{sub2vec}, Graph2Vec \cite{graph2vec}, InfoGraph \cite{infograph}, GraphCL \cite{graphcl}, ADGCL \cite{adgcl}, JOAO \cite{joao}, RGCL \cite{rgcl}, SimGRACE \cite{simgrace}, HTML \cite{html}, DRGCL \cite{drgcl} and CellCLAT \cite{cellclat}. For the semi-supervised setting, we select ten of these unsupervised baselines for comparison.

\subsubsection{Running Environment}

Our method is implemented using PyTorch 1.7. All experiments are conducted on a NVIDIA Tesla V100 GPU, running on Ubuntu 20.04. 

\subsection{Unsupervised Learning}

The results of unsupervised graph-level representation learning for downstream graph classification tasks are summarized in Table \ref{tab:htggcl_unsupervised_learning}. HTG-GCL consistently achieves the best performance across all datasets, with an impressive average rank of 1.0, outperforming all other methods. 
This highlights the effectiveness of HTG-GCL in learning discriminative and stable graph representations.
Furthermore, from the perspective of statistical significance, HTG-GCL achieves strongly or statistically significant improvements 
($p<0.01$ or $p<0.05$) 
over the second-best method on most datasets, as indicated by the asterisks in the table. This demonstrates that the performance gains of HTG-GCL are not only consistent but also statistically reliable.

\subsection{Semi-supervised Learning}

Table \ref{tab:htggcl_semisupervised_learning} presents the results of semi-supervised graph-level representation learning for downstream classification tasks. Our method, HTG-GCL, achieves the best overall performance with an average rank of 1.0 across all datasets, outperforming all other methods. 
This outstanding performance highlights the effectiveness of HTG-GCL in generating high-quality representations during unsupervised pretraining, which can be effectively exploited by downstream classifiers even in low-label regimes through pseudo-labeling. 

\subsection{Ablation Studies}

We conducted ablation studies on different granularities and the MGDC module in unsupervised settings, as shown in Table \ref{tab:ablation study}.
The results clearly indicate that multi-granularity integration outperforms single-granularity topological information, demonstrating the effectiveness of incorporating hierarchical topological information.
Moreover, MGDC shows larger improvement in multi-granularity scenarios, highlighting its ability to reduce redundancy and enhance task-relevant information. 
\ours\ demonstrates outstanding performance across all settings, validating the importance of hierarchical topological granularity integration in improving graph contrastive learning tasks.

\begin{figure}
    \centering
    \includegraphics[width=0.8\linewidth]{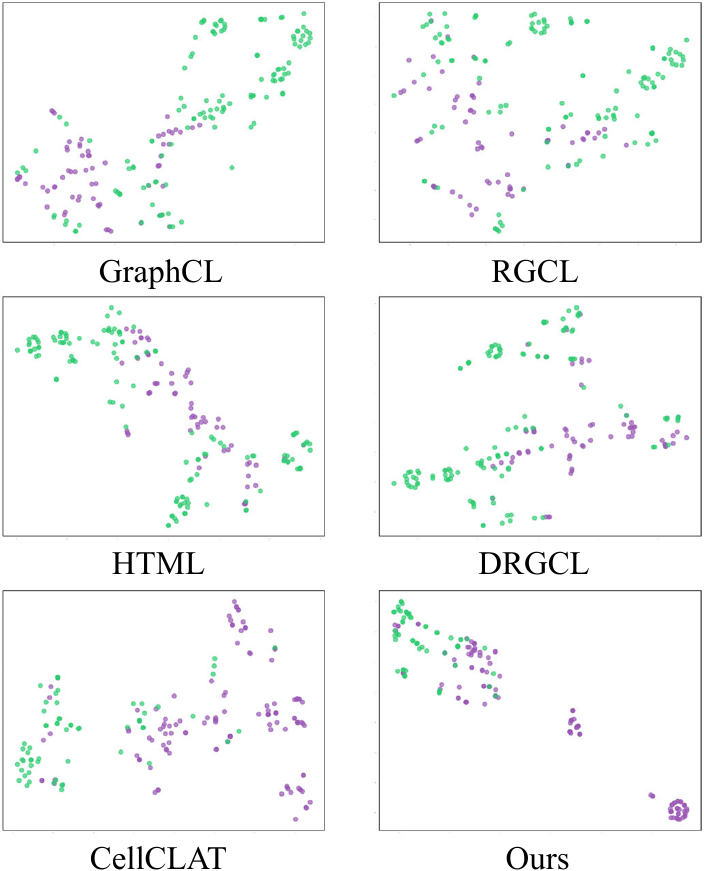}
    \caption{t-SNE visualization of six methods on MUTAG.}
    \label{fig:tsne}
\end{figure}

\subsection{Hyper-parameter Analysis}

\subsubsection{Number of Granularities}

We investigate the impact of the number of granularities from 1 to 4 on model performance. The granularities represent the following combinations of granularities: [6], [6,9], [6,9,12], [6,9,12,15].
Figure \ref{fig:hyper-main}(a) shows the classification accuracy on the NCI1 and NCI109 datasets. The highest accuracy is achieved with 3 granularities, and performance declines when the number of granularities is either lower or higher. The worse performance of more granularitie [6,9,12,15] may be due to the scarcity of rings larger than 12 in the NCI dataset.

\subsubsection{Coefficient $\beta$ for decoupled granularity loss}

Figure \ref{fig:hyper-main}(b) illustrates the classification accuracy on NCI1 and NCI109 dataset across different values of $\beta$ $\in  \{1,0.1,0.01,0.001,0.0001\}$ . We observe that both datasets achieve the best performance when $\beta$ is set to 0.01.

\subsubsection{Batch Size}
Figure \ref{fig:hyper-main}(c) shows the classification accuracy after training for 20 epochs using different batch sizes ranging from 32 to 256 on NCI1 and NCI109 datasets. We observe the performance improves as the batch size increases, with the best results achieved at a batch size of 128.

\subsubsection{lr}
Figure \ref{fig:hyper-main}(d) illustrates the sensitivity of the classification performance to the learning rate, with values ranging from 1 to 0.0001. Both datasets achieve the highest accuracy when the learning rate is set to 0.001.


\subsection{Visualization Results}

In Figure \ref{fig:tsne}, we present the t-SNE visualizations of six different methods on the MUTAG dataset. Our method exhibits better separation between classes, indicating its superior ability to learn discriminative representations compared to the baseline methods.

\subsection{Time Complexity Analysis}

\subsubsection{Multi-granularity Graph-to-Complex Transformation}
For each maximum ring size $m \in \{6,9,12\}$, we attach 2-cells to all induced cycles of length at most $m$, thereby constructing a cellular complex $\{X_{m}(G_i)\}_{m}$. This transformation is a one-time preprocessing step before training. Its complexity is bounded by
$\Theta\left((|E| + |V|N_m)\,\mathrm{polylog}|V|\right)$,
where $N_m$ is the number of induced cycles with length at most $m$ \cite{ferreira2014amortized}. Since HTG-GCL constructs $N_m=3$ granularities in practice, the preprocessing cost grows with the total number of rings considered across granularities (thus positively correlated with $N_m$). 

\begin{table}[t]
    \setlength{\tabcolsep}{4pt}
    \begin{center}
            \begin{tabular}{c|cccc}
                \hline
                \hline
                \text{Dataset} & \text{NCI1} & \text{PROTEINS} & \text{MUTAG} 
                 & \text{IMDB-B} \\
                \hline
                \hline
                \text{Time} & 43s & 14s & 53s & 151s \\
                \hline
                \hline
            \end{tabular}
    \end{center}
    \caption{Transformation time across different datasets.}
    \label{tab:transformation-time}
\end{table}

\begin{table}[t]
    \begin{center}
            \begin{tabular}{c|cccc}
                \hline
                \hline
                \text{Dataset} & \text{GraphCL} & \text{DRGCL} & \text{CellCLAT} & \text{Ours} \\
                \hline
                \hline
                \text{PROTEINS} & 2s & 4s & 6s & 7s \\
                \text{NCI1} & 17s & 18s & 15s & 18s \\
                \hline
                \hline
            \end{tabular}
    \end{center}
    \caption{Training time per epoch across different methods.}
    \label{tab:time-comparison}
\end{table}

\subsubsection{Multi-scale Topological Granularity Decoupled Contrastive Learning}
\textit{Message Passing:}
According to \cite{bodnar2021weisfeiler2}, the computational complexity of the message passing scheme is $\Theta\!\left(\sum_{p=1}^{d} S_p\right)$,
where $B_p$ is the maximum boundary size of a $p$-cell in $X$ and $S_p$ denotes the number of $p$-cells.
Since HTG-GCL processes $N_m$ granularities, the passing cost for a batch of n samples is $\Theta\!\left({N_m}n\sum_{p=1}^{d} S_p\right)$.
\textit{Contrastive Loss Computation:}
HTG-GCL performs contrastive learning in both the common space and granularity-specific decoupled spaces. For a batch of $n$ samples and embedding dimension $d$, the similarity matrix has size $(N_m n) \times (N_m n)$, yielding a per-batch complexity $\Theta\big(N_m^2 n^2 d\big)$.
\textit{Uncertainty-based Weighting Module:}
In each granularity-specific space, we compute clustering distributions $p_{iq}^{g_a}$ using Student’s t-distribution (Eq.~\ref{eq:student-t}) and then map them into Dirichlet parameters (Eq. 8-9). For a batch of $n$ samples with embedding dimension $d$ and $K$ clusters ($K=n$), the per-batch complexity is $\Theta\big((N_m+1)\, n^2 d\big)$. 
Combining the above three parts, the total complexity of HTG-GCL is
$
\Theta\big({N_m}n\sum_{p=1}^{d} S_p\big) 
+ \Theta\big(N_m^2 n^2 d\big)
+ \Theta\big((N_m+1)\, n^2 d\big).
$

\section{Conclusion}

In this paper, we propose HTG-GCL,
a framework that addresses the limitations of fixed-granularity contrastive learning. By transforming graphs into multi-scale ring-based cellular complexes, HTG-GCL constructs diverse topological views to capture fine- and coarse-grained structures. To mitigate conflicts across views, we introduce a decoupled contrastive strategy and assign trustworthiness-based weights to granularity-specific representation spaces. Extensive experiments on benchmark datasets show that HTG-GCL achieves consistent improvements over state-of-the-art methods. This work highlights the importance of hierarchical granularity modeling and opens new directions for adaptive contrastive learning on topological graph structures.

\clearpage

\section*{Acknowledgements}
The authors would like to thank the editors and reviewers for their valuable comments.
This work is supported by the National Natural Science Foundation of China No. 62406313, 2023 Special Research Assistant Grant Project of the Chinese Academy of Sciences.

\bibliography{aaai2026}

\clearpage

\end{document}